\documentclass[journal]{IEEEtran}
\IEEEoverridecommandlockouts
\usepackage{cite}
\usepackage{amsmath,amssymb,amsfonts}
\usepackage{algorithmic}
\usepackage{graphicx}
\usepackage{textcomp}
\usepackage{xcolor}
\usepackage{gensymb}

\def\BibTeX{{\rm B\kern-.05em{\sc i\kern-.025em b}\kern-.08em
    T\kern-.1667em\lower.7ex\hbox{E}\kern-.125emX}}
\begin{document}

\title{Fetal Brain Imaging: A Composite Neural Network Approach for Keyframe Detection in Ultrasound Videos
\thanks{© 2023 IEEE. Personal use of this material is permitted. Permission from IEEE must be obtained for all other uses, in any current or future media, including reprinting/republishing this material for advertising or promotional purposes, creating new collective works, for resale or redistribution to servers or lists, or reuse of any copyrighted component of this work in other works.

The final published version is available in IEEE Xplore at: 10.1109/PAEE59932.2023.10244374}
}

\author{\IEEEauthorblockN{
1\textsuperscript{st} Aleksander Zamojski}
\IEEEauthorblockA{\textit{Warsaw Universitu of Technology} \\
Warsaw, Poland \\
}
\and
\IEEEauthorblockN{2\textsuperscript{nd} Kacper Jarczak}
\IEEEauthorblockA{\textit{Warsaw Universitu of Technology} \\
Warsaw, Poland \\
}
\and
\IEEEauthorblockN{3\textsuperscript{rd} Radosław Roszczyk}
\IEEEauthorblockA{\textit{Warsaw University of Technology} \\
Warsaw, Poland \\
radoslaw.roszczyk@pw.edu.pl}
}

\maketitle

\begin{abstract}
This article presents a novel approach to keyframe detection in ultrasound videos, with a particular focus on fetal brain imaging. The proposed model is a composite neural network architecture that combines a Convolutional Neural Network (CNN) with a Recurrent Neural Network (RNN). The CNN extracts spatial features from individual video frames, while the RNN captures temporal dependencies between consecutive frames within each video sequence.

The proposed model may improve the efficiency and accuracy of fetal brain ultrasound analysis, thereby supporting earlier detection, diagnosis, and treatment planning for selected fetal brain conditions. 
\end{abstract}


\section{Introduction}
Ultrasonography is a widely used tool for examinations used in fetal diagnosis. The main advantages of this examination are its relatively low cost, high mobility and the ability to examine in real-time. For example, the prenatal weight of the baby can be estimated from ultrasound measurements of head circumference, bi coronal diameter, abdominal circumference and femur length \cite{Salomon2011}.

This examination is performed according to a specific protocol using a manual scan. Based on this, the doctor selects a specific plane to make the appropriate biomedical measurements. For some conditions, accurate data acquisition and the precise selection of ultrasound planes that clearly and representative represent critical fetal anatomical structures can be difficult \cite{Gangarosa2005}.

Ultrasonography also allows the diagnosis of congenital heart defects and arrhythmias. It is also possible to assess cardiac and cardiovascular function and diagnose diseases such as heart failure, twin transfusion syndrome, detection of pulmonary masses and vascular tumours \cite{Donofrio2014}.

Furthermore, it is possible to detect genetic syndromes in fetuses with sonographic abnormalities but typical microarray results \cite{Drury2015, Lord2019}.

\section{Related Work}
The problem of keyframe detection is fundamental from the point of view of medical diagnosis. The extraction of relevant information from video sequences makes it possible to automate the process of assisting medical diagnosis. Several researchers are working on video analysis and methods that allow the detection of relevant individual frames of the video. For example, \cite{Liu2003} proposed a triangular perceived motion energy (PME) model for modelling motion patterns in video and a scheme for extracting key frames based on this model. The frames at the turning point of acceleration and deceleration of motion are selected as keyframes. Aoki proposed a different approach \cite{Aoki1996}. His solution used a method to reduce redundancy in keyframes by detecting repetitions using chromatic histograms and differences in local average luminance values. Ott proposed a similar solution \cite{Ott2007}. It uses a keyframe selection method based on colour histogram filtering.
The mechanisms presented are pretty effective for video recordings but do not work well for ultrasound recordings. A different approach should be used for this purpose. In the paper \cite{Singh2022} an automated method for detecting diagnostically relevant frames containing the optic nerve sheath from video ultrasound of the eye using deep learning is presented. This solution highlights the specific elements of ultrasound and the problems associated with identifying key frames. Salomon, in his paper {Salomon}, presents an entire procedure with guidelines for performing routine mid-trimester fetal ultrasounds. The mechanisms presented allow the identification of medically appropriate images. A completely different approach is presented in the Bamber article{Bamber}. It discusses the basic principles and technology of ultrasound elastography, a method that can be used to select keyframes in ultrasound imaging. It mentions that the effective use of elastography requires knowledge of fundamental physics and technology, which is hampered by the complexity of the subject and the variety of technologies available. However, this complexity presents opportunities the research community is enthusiastic to exploit. As a result, elastography technology is likely to undergo significant development beyond the currently available techniques. Significant improvements in image quality, ease of use, quantification and range of measurable tissue features can be expected.

\section{Method}

\begin{figure*}
\begin{center}
\includegraphics[width=\linewidth]{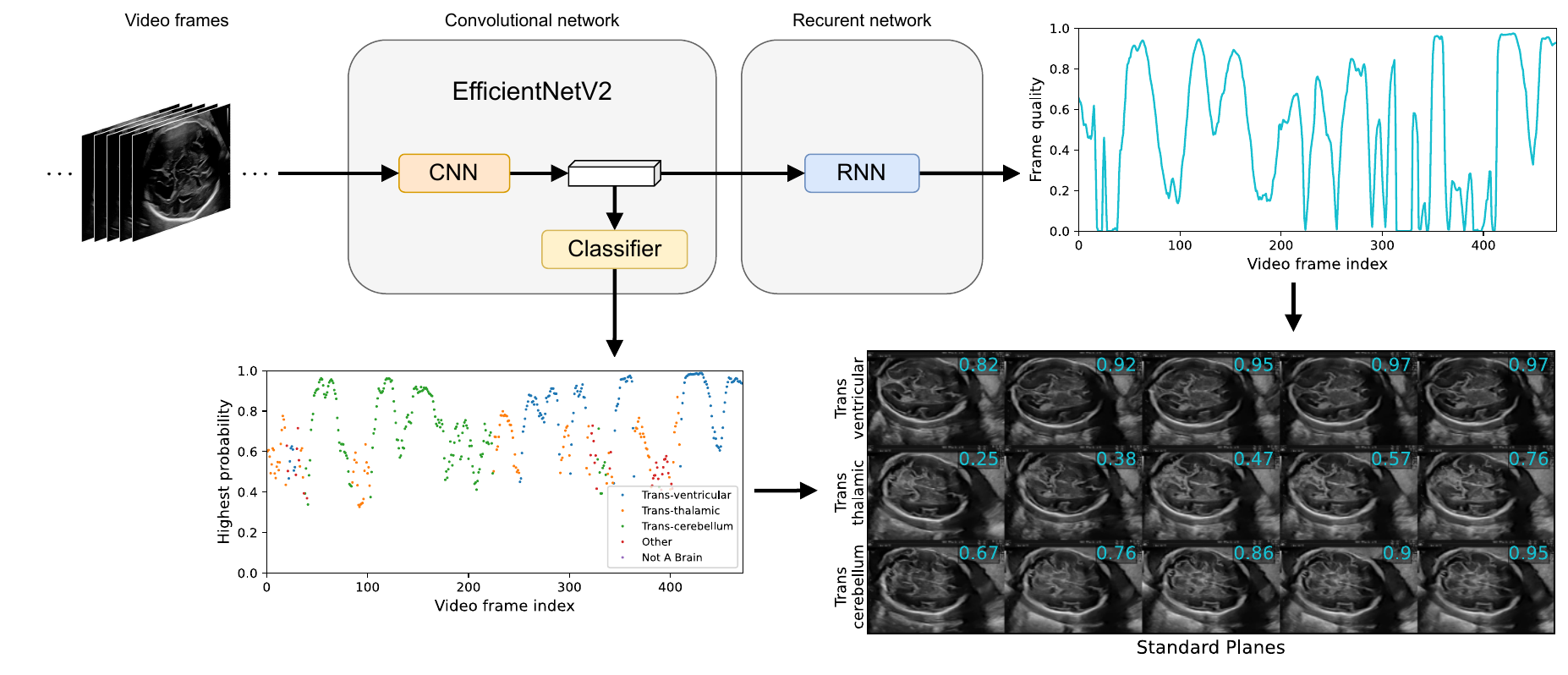}
\end{center}
   \caption{Overview of proposed composite neural network model.}
\label{fig:CompositeNeuralNetwork}
\end{figure*}
The following section will outline the approach used to train the proposed composite model for detecting the best frames of the ultrasound video on which the brain planes were located. Due to the heterogeneity of the problem and the lack of a clearly defined metric for frame quality, an approach based on a composite neural network was chosen.

The proposed network consists of two main units that are separate sub-networks.
First, the Convolutional Neural Network (CNN) is used to discover features within individual video frames. The second component, the Recurrent Neural Network (RNN), is responsible for discovering relations between consecutive frames within a single video.

The first step of the described approach is to split the video into individual frames, which are then processed by CNN. The final, fully connected layer outputs the probability of belonging to five groups which are: the three most commonly used brain planes (Trans-ventricular, Trans-thalamic, Trans-cerebellar), parts of the brain different than selected planes (brain(other)) and images that do not represent the brain (not a brain).

The output of the penultimate CNN (dense output) is then passed to the input of the RNN. The recursive network processes the features of the subsequent video frames and produces a quality value for each image.

The composite neural network was learned in a stepwise fashion. The CNN classifier was the first to be learned. Subsequently, a frame quality metric for ultrasound videos was developed using the classifier's output. Finally, using the output features from the CNN and the calculated quality measure, an RNN was trained.

\subsection{CNN}


\subsubsection{Data}

\begin{table}
\footnotesize 
\addtolength{\tabcolsep}{-1pt} 
\begin{center}
\begin{tabular}{|ll|r|r|r|}
\hline
\multicolumn{2}{|l|}{\textbf{Anatomical planes}} & \textbf{N. patients} & \textbf{N. images} & \textbf{N. train images} \\
\hline
\multicolumn{2}{|l|}{Fetal brain}   & 1,082  & 3,092 & 1556 \\
  & Trans-ventricular               & 446    & 597   & 231 \\
  & Trans-thalamic                  & 909    & 1,638 & 873 \\
  & Trans-cerebellum                & 575    & 714   & 375 \\
  & Other                           & 575    & 143   & 77 \\
\multicolumn{2}{|l|}{Not a brain}   & 1,731  & 9,308  & 5,509 \\
\hline
\end{tabular}
\addtolength{\tabcolsep}{1pt} 
\end{center}
\caption{Maternal-fetal US dataset statistics: anatomical planes labeled, number of patients, number of images, number of images in training set~\cite{Xavier2020}}
\label{tab:StandardPlanesDataset}
\end{table}

A publicly available dataset was used to learn the CNN classifier. It consists of 12,400 labelled fetal US images from 1,792 patients, of which 3,092 are brain images \cite{Xavier2020}. Fetal brain images are divided into four groups representing the three most popular brain planes and the "other" section where neither plane was assigned. The exact numbers of images in each group are presented in the table~\ref{tab:StandardPlanesDataset}.
\\
\subsubsection{Network architecture}

As the architecture of the CNN was not the subject of research per se, it was decided to select the most suitable solution from those already in use. The following solutions were investigated: densenet169, mobilenet\_v3\_small, mobilenet\_v3\_large, efficientnet\_v2\_s, efficientnet\_v2\_m, resnet18, resnet34, resnet50, resnet101, resnet152, resnext50\_32x4d, resnext101\_32x8d, resnext101\_64x4d
A summary of the tests is presented in the table~\ref{tab:StandardPlanesDataset}
Due to the high accuracy scores and the low processing time of the test set, the EfficientNetV2 (small) architecture was finally selected.
\\
\subsubsection{Training}


Before the learning process started, the part of the dataset not used for testing was split into a training subset(~80\%) and a validation subset (~20\%). Additionally, the division was imposed with the requirement that all images of a single patient be placed in only one of the sets.

Due to the unbalanced number of images in each class, under-sampling was used before each epoch in such a way that the 'not a brain' class had only 500 elements. In addition, in order to bring information about more features into the model, a new random subset of the 'not a brain' class was created for each epoch. The remaining classes were used in entirety. 

Network learning was constrained by a fixed upper limit on the number of epochs of 50. Early stopping set at 10 epochs was also used. The learning rate was 5e-04 and was halved each time the error stopped decreasing. A batch size of 32 was used.

In order to reduce overfitting, data augmentation based on affine transformations was used. New images were created by scaling, horizontal and vertical flipping, translation by a vector, and rotation of images from the base set.

Validation was carried out four times within a single batch. Tests were carried out on network weights obtained from the learning stage, in which the network achieved the highest validation accuracy.

\subsection{Frame quality value}

\begin{figure}[t]
\begin{center}
\includegraphics[width=0.9\linewidth]{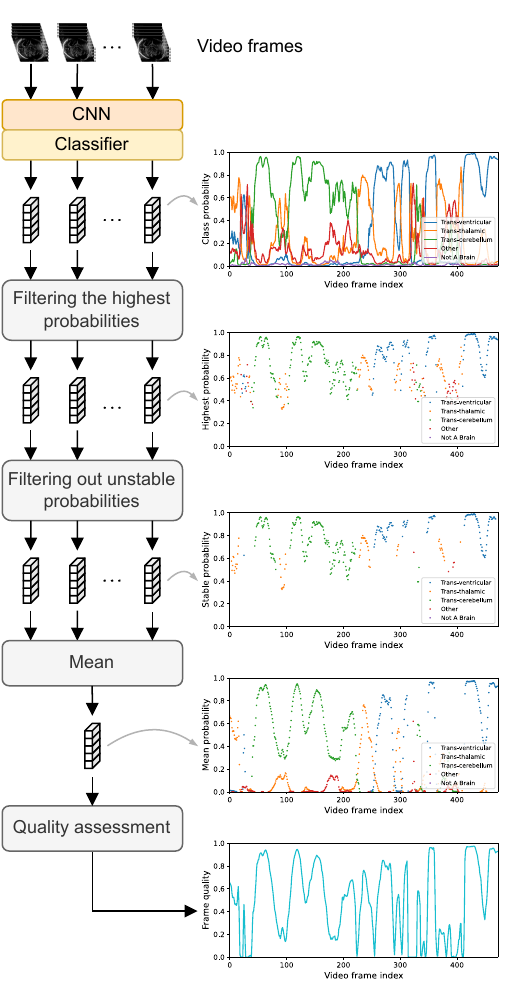}
\end{center}
   \caption{Frame quality computation.}
\label{fig:FrameQualityComputation}
\end{figure}

To determine Frame quality value, a metric based on the response of the CNN classifier to US images was developed. A value in the range 0-1, where scores closer to unity indicate higher quality frames, is derived from the following steps:

As a result of the classification, all video frames receive a set of probabilities of belonging to considered classes.

In the next step, within each frame, the class with the highest probability is marked, while the probabilities of the other classes are assigned zero

In further step, unstable frames are filtered out. Frame "i" is considered unstable if for any of the frames [i - window\_size, i +window\_size] in the previous step, a class other than the class of frame "i" is selected.

To increase the solution's reliability, the previous steps are repeated for the video modified by affine transformations (each video frame is transformed in the same way). The transformations used are: horizontal flip, rotation (+-15\degree, +-10\degree, +-5\degree), translation +-10\%, scale (5\%, 10\%, 15\%, 20\%)). As a result, each frame has as many sets of probabilities as the number of transformed videos entered. The probabilities for a frame are aggregated by calculating the average probability of each class.

In the last step, the quality value of the frame is determined by subtracting the probabilities of the other classes from the value of the highest probability. A diagram of the frame quality computation is provided in Figure \ref{fig:FrameQualityComputation}.

\subsection{RNN}

The role of the RNN is to predict frame quality values based on the already processed frames.
\\
\subsubsection{Data}

\begin{table}
\footnotesize 
\begin{center}
\begin{tabular}{|l|r|r|r|r|}
\hline
 & Min & Max & Mean & Std \\
\hline
 Duration (second) & 4 & 50 & 16.42 & 7.85 \\
 Frames per second & 22 & 55 & 29.67 & 3.63 \\
\hline\hline
 Number of videos & \multicolumn{4}{|l|}{130} \\
\hline
\end{tabular}
\end{center}
\caption{Maternal-fetal US videos statistics}
\label{tab:USVideoDataset}
\end{table}

Dataset used for RNN training was obtained by agreement with a local medical facility and consists of 130 short (4-50s) US pregnancy screening videos which statistics has been shown in the table~\ref{tab:USVideoDataset}. Received fragments show mainly the fetal brain, but some frames do not contain it.
\\
\subsubsection{Network architecture}


The proposed Recurrent Neural Network (RNN) architecture uses the Gated Recurrent Unit (GRU) as the primary recurrent element, and it is structured as follows: 
\begin{itemize}
    \item Input Layer: The RNN accepts an input size of 1280.
    \item First GRU Layer: The input data flows into the first GRU layer
    \item First Dropout Layer(0.1):  Immediately following the first GRU layer, the first dropout layer is applied. The dropout rate is set to 0.1, meaning 10\% of the input units to the layer are dropped or "turned off" during training, which helps prevent overfitting.
    \item Second GRU Layer: The output from the first dropout layer is then passed to the second GRU layer. This layer is intended to capture further long-term dependencies based on the input from the previous layer.
    \item Second Dropout Layer (0.2): A second dropout layer is applied with a dropout rate of 0.2, or 20\%. This further helps in reducing overfitting by preventing complex co-adaptations on the training data.
    \item  Fully Connected Layer (1): The output from the second dropout layer is passed to a fully connected (dense) layer.
    \item  Sigmoid Activation Function: The output of the fully connected layer is passed through a sigmoid activation function
\end{itemize}

Thus, the architecture is designed to process the input through successive layers of GRU, Dropout, and finally a fully connected layer with a sigmoid activation function, thereby making a binary decision on the input data.
\\
\subsubsection{Training}


Before the learning process started, the part of the dataset not used for testing was split into a training subset(~90\%) and a validation subset (~10\%).

Network learning was constrained by a fixed upper limit on the number of epochs of 60. Early stopping set at 20 epochs was also used. The learning rate was 5e-04. A weight decay of value 1e-05 was used.



\section{Result}
The presented research delineates the development and implementation of a composite neural network model designed for the detection of key frames within ultrasound videos, focusing predominantly on imaging of the fetal brain. The model features two main units, a Convolutional Neural Network (CNN) and a Recurrent Neural Network (RNN), each with its unique role within the process.

The CNN unit, trained via a public dataset encompassing 12,400 labeled fetal ultrasound images derived from 1,792 patients, was optimized to discern the three most frequently utilized brain planes (Trans-ventricular, Trans-thalamic, Trans-cerebellar), segments of the brain not corresponding to the selected planes (brain(other)), and images devoid of any brain representation (not a brain). For the implementation of the CNN, the EfficientNetV2 (small) architecture was chosen, owing to its superior accuracy and efficient processing times when evaluated against the test set.

In addition to the CNN's capabilities, a frame quality metric was developed, which utilized the classifier's output. This allowed for a further layer of analysis, where the RNN was trained using the output features obtained from the CNN along with the calculated quality measure. The RNN, thus, serves the purpose of predicting the quality value of the frames based on previously processed ones.

The present study delves into the keyframe detection challenge from a medical diagnostic standpoint, establishing its crucial role in automating the process of medical diagnostic assistance. Various methodologies for keyframe detection within video analysis have been critically evaluated, such as perceived motion energy models, chromatic histogram-based approaches, color histogram filtering, and ultrasound elastography. Yet, it was determined that these existing strategies do not translate effectively into ultrasound recordings, underscoring the need for innovative solutions like the one proposed in this study.

\section{Conclusion}
This research represents a significant stride forward in the realm of keyframe detection in ultrasound videos, specifically those related to fetal brain imaging. By leveraging a composite neural network model, constituted of a Convolutional Neural Network (CNN) and a Recurrent Neural Network (RNN), we have illuminated a novel pathway that holds promise for advancing the automation of medical diagnostic processes.

The CNN, instrumental in distinguishing features within individual video frames, in combination with the RNN, which excels in identifying correlations between successive frames within a single video, constitutes a robust mechanism that is poised to address the challenges associated with fetal brain imaging. By utilizing a publicly available dataset for training the CNN and then using the output features from the CNN to train the RNN, a model was created that effectively calculates and predicts frame quality value.

The potential impact of these findings is profound in the field of medical imaging, specifically within the context of prenatal diagnostics. With the capability of enhancing the accuracy and efficiency of fetal brain imaging, this proposed model paves the way towards improving early detection and intervention for various fetal conditions, hence leading to more effective treatments and better patient outcomes.

However, it is crucial to underscore that further studies and validations are required to verify the model's performance within real-world clinical environments. Despite the promising initial findings, broader implementation and analysis remain essential to thoroughly gauge the model's overall effectiveness.

Ultimately, the study sets the stage for future research to explore how such advanced technologies can continue to be optimized and adapted within the ever-evolving field of ultrasound imaging. The success of the proposed model also hints at the potential for its application to other areas of medical imaging, opening new avenues for future exploration and innovation. Moreover, the study paves the way for future exploration into the untapped potential of the model within ultrasound imaging.

\bibliographystyle{ieeetr}
\bibliography{export}

@article{Gangarosa2005,
   author = {Lisa M. Gangarosa},
   doi = {10.1053/j.gastro.2005.08.038},
   issn = {00165085},
   issue = {4},
   journal = {Gastroenterology},
   month = {10},
   pages = {1357},
   title = {The Practice of Ultrasound: A Step-by-Step Guide to Abdominal Scanning},
   volume = {129},
   url = {https://linkinghub.elsevier.com/retrieve/pii/S0016508505017531},
   year = {2005},
}

@inproceedings{Ott2007,
   author = {L. Ott and P. Lambert and B. Ionescu and D. Coquin},
   doi = {10.1109/ICIAPW.2007.12},
   isbn = {0-7695-2921-6},
   journal = {14th International Conference of Image Analysis and Processing - Workshops (ICIAPW 2007)},
   month = {9},
   pages = {206-211},
   publisher = {IEEE},
   title = {Animation Movie Abstraction: Key Frame Adaptative Selection Based on Color Histogram Filtering},
   url = {http://ieeexplore.ieee.org/document/4427502/},
   year = {2007},
}

@inproceedings{Aoki1996,
   author = {Hisashi Aoki and Shigeyoshi Shimotsuji and Osamu Hori},
   city = {New York, New York, USA},
   doi = {10.1145/244130.244135},
   isbn = {0897918711},
   journal = {Proceedings of the fourth ACM international conference on Multimedia  - MULTIMEDIA '96},
   pages = {1-10},
   publisher = {ACM Press},
   title = {A shot classification method of selecting effective key-frames for video browsing},
   url = {http://portal.acm.org/citation.cfm?doid=244130.244135},
   year = {1996},
}

@article{Liu2003,
   author = {Tianming Liu and Hong-Jiang Zhang and Feihu Qi},
   doi = {10.1109/TCSVT.2003.816521},
   issn = {1051-8215},
   issue = {10},
   journal = {IEEE Transactions on Circuits and Systems for Video Technology},
   month = {10},
   pages = {1006-1013},
   title = {A novel video key-frame-extraction algorithm based on perceived motion energy model},
   volume = {13},
   url = {http://ieeexplore.ieee.org/document/1234141/},
   year = {2003},
}

@article{Lord2019,
   author = {Jenny Lord and Dominic J McMullan and Ruth Y Eberhardt and Gabriele Rinck and Susan J Hamilton and Elizabeth Quinlan-Jones and Elena Prigmore and Rebecca Keelagher and Sunayna K Best and Georgina K Carey and Rhiannon Mellis and Sarah Robart and Ian R Berry and Kate E Chandler and Deirdre Cilliers and Lara Cresswell and Sandra L Edwards and Carol Gardiner},
   doi = {10.1016/S0140-6736(18)31940-8},
   issn = {01406736},
   issue = {10173},
   journal = {The Lancet},
   month = {2},
   pages = {747-757},
   pmid = {30712880},
   publisher = {Lancet Publishing Group},
   title = {Prenatal exome sequencing analysis in fetal structural anomalies detected by ultrasonography (PAGE): a cohort study},
   volume = {393},
   url = {https://linkinghub.elsevier.com/retrieve/pii/S0140673618319408},
   year = {2019},
}

@article{Drury2015,
   author = {Suzanne Drury and Hywel Williams and Natalie Trump and Christopher Boustred and Nicholas Lench and Richard H. Scott and Lyn S. Chitty},
   doi = {10.1002/pd.4675},
   issn = {01973851},
   issue = {10},
   journal = {Prenatal Diagnosis},
   month = {10},
   pages = {1010-1017},
   pmid = {26275891},
   publisher = {John Wiley and Sons Ltd},
   title = {Exome sequencing for prenatal diagnosis of fetuses with sonographic abnormalities},
   volume = {35},
   url = {https://onlinelibrary.wiley.com/doi/10.1002/pd.4675},
   year = {2015},
}

@article{Salomon2011,
   author = {L. J. Salomon and Z. Alfirevic and V. Berghella and C. Bilardo and E. Hernandez‐Andrade and S. L. Johnsen and K. Kalache and K.‐Y. Leung and G. Malinger and H. Munoz and F. Prefumo and A. Toi and W. Lee},
   doi = {10.1002/uog.8831},
   issn = {0960-7692},
   issue = {1},
   journal = {Ultrasound in Obstetrics \& Gynecology},
   month = {1},
   pages = {116-126},
   pmid = {20842655},
   title = {Practice guidelines for performance of the routine mid‐trimester fetal ultrasound scan},
   volume = {37},
   url = {https://onlinelibrary.wiley.com/doi/10.1002/uog.8831},
   year = {2011},
}

@article{Donofrio2014,
   author = {Mary T. Donofrio and Anita J. Moon-Grady and Lisa K. Hornberger and Joshua A. Copel and Mark S. Sklansky and Alfred Abuhamad and Bettina F. Cuneo and James C. Huhta and Richard A. Jonas and Anita Krishnan and Stephanie Lacey and Wesley Lee and Erik C. Michelfelder and Gwen R. Rempel and Norman H. Silverman and Thomas L. Spray and Janette F. Strasburger and Wayne Tworetzky and Jack Rychik},
   doi = {10.1161/01.cir.0000437597.44550.5d},
   issn = {0009-7322},
   issue = {21},
   journal = {Circulation},
   month = {5},
   pages = {2183-2242},
   pmid = {24763516},
   publisher = {Lippincott Williams and Wilkins},
   title = {Diagnosis and Treatment of Fetal Cardiac Disease},
   volume = {129},
   url = {https://www.ahajournals.org/doi/10.1161/01.cir.0000437597.44550.5d},
   year = {2014},
}

@article{Singh2022,
   author = {Maninder Singh and Basant Kumar and Deepak Agrawal},
   doi = {10.1007/s11517-022-02680-3},
   issn = {0140-0118},
   issue = {12},
   journal = {Medical \& Biological Engineering \& Computing},
   month = {12},
   pages = {3397-3417},
   publisher = {Springer Science and Business Media Deutschland GmbH},
   title = {Good view frames from ultrasonography (USG) video containing ONS diameter using state-of-the-art deep learning architectures},
   volume = {60},
   url = {https://link.springer.com/10.1007/s11517-022-02680-3},
   year = {2022},
}

@article{Xavier2020,
   author = {Xavier P. Burgos-Artizzu and David Coronado-Gutiérrez and Brenda Valenzuela-Alcaraz and Elisenda Bonet-Carne and Elisenda Eixarch and Fatima Crispi and Eduard Gratacós},
   doi = {10.1038/s41598-020-67076-5},
   issn = {2045-2322},
   issue = {1},
   journal = {Scientific Reports},
   month = {6},
   pages = {10200},
   title = {Evaluation of deep convolutional neural networks for automatic classification of common maternal fetal ultrasound planes},
   volume = {10},
   url = {https://www.nature.com/articles/s41598-020-67076-5},
   year = {2020},
}

\end{document}